\documentclass[journal]{IEEEtranTIE}
\usepackage{graphicx}
\usepackage{cite}
\usepackage{picinpar}
\usepackage{amsmath}
\usepackage{url}
\usepackage{flushend}
\usepackage[latin1]{inputenc}
\usepackage{colortbl}
\usepackage{soul}
\usepackage{multirow}
\usepackage{pifont}
\usepackage{color}
\usepackage{alltt}
\usepackage[hidelinks]{hyperref}
\usepackage{enumerate}
\usepackage{siunitx}
\usepackage{breakurl}
\usepackage{epstopdf}
\usepackage{pbox}
\usepackage{amssymb}
\usepackage{multirow}
\usepackage[table,xcdraw]{xcolor}

\begin{document}
\title{Virtual-Tube-Based Cooperative Transport Control for Multi-UAV Systems in Constrained Environments}

\author{
	\vskip 1em
	
	Runxiao Liu,
	Pengda Mao, 
	Xiangli Le,
	Shuang Gu, 
	Yapeng Chen,\\
	and Quan Quan$^{*}$ \emph{Senior Member,IEEE}

	\thanks{
	
		
		Runxiao Liu, Pengda Mao, Xiangli Le, Shuang Gu, Yapeng Chen, Kai-Yuan Cai, Quan Quan (Corresponding Author) are with School of Automation Science and Electrical Engineering, Beihang University, Beijing, 100191, P.R. China (runxiao\_liu@buaa.edu.cn, qq\_buaa@buaa.edu.cn). 
		
	}
}

\maketitle
	
\begin{abstract}
This paper proposes a novel control framework for cooperative transportation of cable-suspended loads by multiple unmanned aerial vehicles (UAVs) operating in constrained environments. Leveraging virtual tube theory and principles from dissipative systems theory, the framework facilitates efficient multi-UAV collaboration for navigating obstacle-rich areas. The proposed framework offers several key advantages. (1) It achieves tension distribution and coordinated transportation within the UAV-cable-load system with low computational overhead, dynamically adapting UAV configurations based on obstacle layouts to facilitate efficient navigation. (2) By integrating dissipative systems theory, the framework ensures high stability and robustness, essential for complex multi-UAV operations. The effectiveness of the proposed approach is validated through extensive simulations, demonstrating its scalability for large-scale multi-UAV systems. Furthermore, the method is experimentally validated in outdoor scenarios, showcasing its practical feasibility and robustness under real-world conditions.
\end{abstract}

\begin{IEEEkeywords}
	ADA Control, Cooperative Transport, Distributed Control, Dissipative System, Virtual Tube
\end{IEEEkeywords}

\markboth{IEEE TRANSACTIONS ON INDUSTRIAL ELECTRONICS}%
{}

\definecolor{limegreen}{rgb}{0.2, 0.8, 0.2}
\definecolor{forestgreen}{rgb}{0.13, 0.55, 0.13}
\definecolor{greenhtml}{rgb}{0.0, 0.5, 0.0}

\section{Introduction}

\IEEEPARstart{T}{he} advancements in sensing and control technologies for unmanned aerial vehicles (UAVs) \cite{8333748,10350055} have catalyzed the emergence of UAV-based load transportation as a prominent area of research \cite{menouar2017uav,al2020uav}. UAV transportation offers distinct advantages over traditional ground-based systems, including enhanced flexibility, higher speed, and the ability to traverse complex or inaccessible terrains \cite{6873329}. Consequently, UAVs hold significant potential for various applications, particularly in emergency response scenarios and the rapid transportation of materials \cite{bernard2011autonomous,thiels2015use}.

In practice, the weight of the load can vary significantly depending on the specific requirements, particularly when dealing with heavy loads \cite{hegde2022multi}. Such variability often exceeds the capacity of a single UAV, rendering conventional UAV designs insufficient. Two primary approaches have been proposed to address this challenge: (1) designing UAVs with greater load capacities \cite{ong2019design} or (2) employing multiple UAVs in a cooperative manner to share the load. While the first approach requires substantial investment in terms of time and resources for developing new UAV models, the second approach multi-UAVs cooperative transportation offers notable advantages, including cost-effectiveness, robustness, and adaptability to a range of loads and environmental conditions \cite{wu2023cooperative,zhang2023formation,su2023robust,rao2023path,zhang2021self,liu2024coordinated}.

\begin{figure}[!h]
	\centerline{\includegraphics[width=0.5\textwidth]{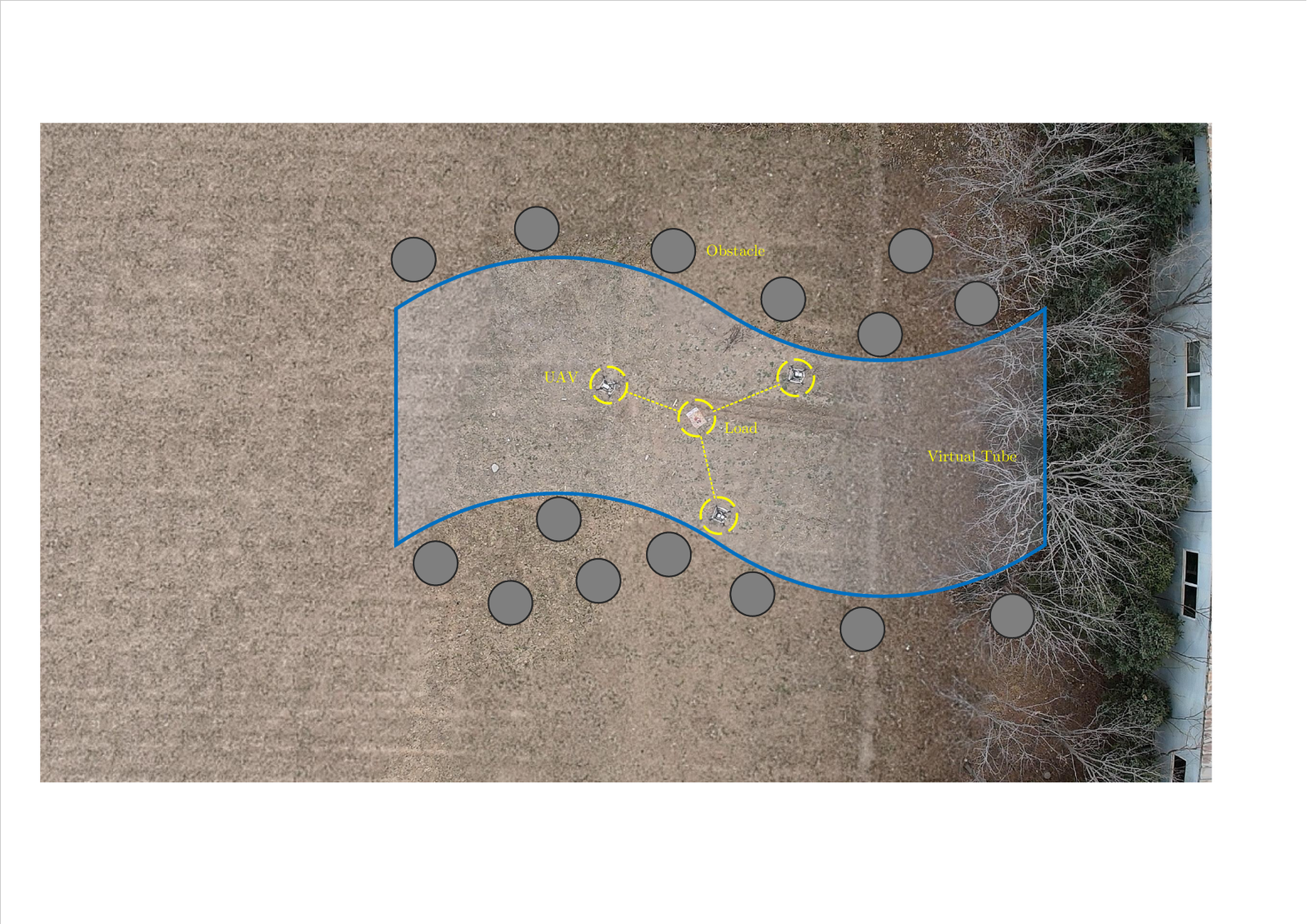}}
	\caption{Outdoor experiments of cooperative transport by multiple UAVs with virtual tube}
	\label{fig:exp1_1}
\end{figure}

Existing approaches to multi-UAV cooperative transport can be broadly categorized into three methodological frameworks: formation-based control \cite{su2023robust,zhang2021self,liu2024coordinated}, planning-based control \cite{zhang2023formation,rao2023path,loianno2017cooperative,sreenath2013dynamics}, and admittance control \cite{wu2023cooperative,geng2020cooperative}. Formation-based control is computationally efficient but lacks mechanisms for distributing cable tension among the UAVs. This limitation becomes particularly pronounced in complex scenarios, such as multi-obstacle or narrow passage environments, where frequent manual reconfiguration of the formation is required to achieve collision avoidance and ensure traversal. Planning-based approaches, on the other hand, enable precise cable tension allocation and high control accuracy when the system is accurately modeled. However, these methods face challenges such as high computational complexity, particularly in large-scale multi-UAV systems, where solving optimization problems becomes intractable. The admittance control approach relies on the active movement of a designated leader UAV, with follower UAVs adjusting their motion based on tension estimations. This can lead to imbalances in load distribution, with the leader operating near its load limit while some follower UAVs experience slack cables. Such slack-tension switching can jeopardize the overall stability and safety of the system during transportation.

Despite notable progress in multi-UAV cooperative transport, most existing research is limited to controlled indoor environments equipped with motion capture systems for precise real-time localization \cite{wu2023cooperative,zhang2023formation,su2023robust,rao2023path,wang2024robust}. These studies often fail to address the complex disturbances and environmental uncertainties inherent to outdoor applications. Furthermore, experimental loads are typically lightweight, often below the capacity of a single UAV \cite{wu2023cooperative,zhang2023formation,su2023robust}, which undermines the primary motivation for cooperative transportation to extend load capacity. Therefore, the development of a computationally efficient, high-load, and adaptable multi-UAV cooperative transport method capable of operating in dynamic and complex environments remains a critical and pressing challenge.

\begin{figure}[!htbp]
	\centerline{\includegraphics[width=0.25\textwidth]{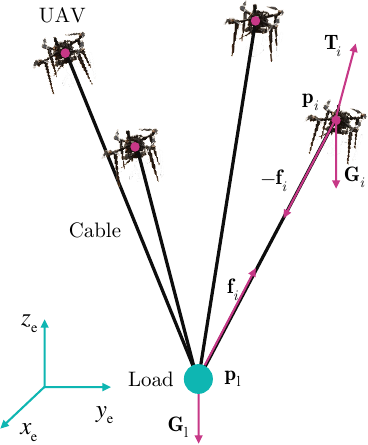}}
	\caption{The UAV-cable-load system}
	\label{fig:UAVsys}
\end{figure}

This paper proposes a cooperative transport control method of multi-UAV in constrained environment based on virtual tube theory and dissipative systems theory. 
It is worth noting that this paper does not merely apply the virtual tube theory to the cooperative transportation. In the virtual tube theory, under the influence of repulsive artificial potential fields, the formation of the UAVs would become increasingly dispersed, preventing the UAVs from achieving a reasonable force distribution for cooperative transportation. However, by integrating the virtual tube theory with dissipative systems theory, the novel control method proposed in this paper enables UAVs to maintain a more reasonable force distribution formation, automatically adjusting according to the width of the virtual tube. This work is related to the emerging research direction of autonomous, dependable, and affordable control \cite{cai2021s,ke2023uniform,ren2024data}.
The contributions of this paper are summarized into three aspects.
\begin{enumerate}[1)]
	\item By incorporating the dissipative systems theory into the virtual tube control method, a cooperative transport control method of multiple UAVs in constrained environment is designed. Particularly, we extend the theory of dissipative systems, which is applicable to second-order systems, to first-order systems in the velocity control mode and design the corresponding cooperative transport controller.
	\item The stability of the proposed cooperative transport controller for first-order systems based on dissipative systems is proved. Meanwhile, an equilibrium point with special significance for cooperative transport is given. Furthermore, the convergence of the proposed cooperative transport control controller in constrained environment is analyzed. 
	\item Simulations of up to ten UAVs cooperative transport in constrained environment are conducted to validate the performance of the control method and its applicability to large-scale UAV cooperative transport. Further, we design a real cooperative transport system and complete the experiments in an outdoor environment to validate the practicality of our algorithm. 
\end{enumerate}

\section{Preliminary and Problem Formulation}
\subsection{Cooperative transport model of the UAV-cable-load system}

In the cooperative transport scene, the UAV-cable-load system consists of $n$ UAVs and one load in $\mathbb{R}^3$. 
Each UAV is connected with the load by a cable, as shown in Fig. \ref{fig:UAVsys}.
The dynamic of a UAV in the system is described as a mass point model:

\begin{equation}\label{eq:1-1}
	\left\{ \begin{array}{l}
		{{{\bf{\dot p}}}_i} = {{\bf{\dot v}}_{ i}}\\
		{{{\bf{\dot v}}}_i} = \frac{1}{{{m_i}}}\left( {{{\bf{T}}_i} + {{\bf{G}}_i} - {{\bf{f}}_i}} \right) 
	\end{array} \right. 
\end{equation}
where ${{{\bf p}}_i} \in \mathbb{R}^3$ and ${{{\bf v}}_i} \in \mathbb{R}^3$ are the position and velocity of the $i$th UAV, 
${m_i}$ is the mass of the $i$th UAV,
${\bf{T}}_i,{\bf{G}}_i,{\bf{f}}_i$ are the propulsive force input, the gravity and the cable force to the $i$th UAV, i = 1,2,...,n.

The dynamic of the load in the system is described as a double integrator model as well:
\begin{equation}\label{eq:1-2}
	\left\{ \begin{array}{l}
		{{{\bf{\dot p}}}_{\rm{l}}} = {{\bf{v}}_{\rm{l}}}\\
		{{{\bf{\dot v}}}_{\rm{l}}} = {{\bf{G}}_{\rm{l}}} + {{\bf{f}}_{\rm{l}}} 
	\end{array} \right. 
\end{equation}
where ${{{\bf p}}_{\rm{l}}} \in \mathbb{R}^3$ and ${{{\bf v}}_{\rm{l}}} \in \mathbb{R}^3$ are the position and velocity of the load, 
${m_{\rm{l}}}$ is the mass of the load, ${{\bf{f}}_{\rm{l}}}$ is the cables force to the load.
The relationship between ${{\bf{f}}_{\rm{l}}}$ and ${\bf{f}}_i$ is
\begin{equation}\label{eq:1-3}
	{{\bf{f}}_{\rm{l}}} = \sum\limits_{i = 1}^n {{{\bf{f}}_i}},i = 1,2,...,n .
\end{equation}

The model of the cable force is 
\begin{equation}\label{eq:1-4}
	\left\{ \begin{aligned}
		{{\bf{f}}_i} =& \max \left( {\min \left( {{A_i},{f_{\max ,i}}} \right),0} \right)\left( { - {\bf{t}}_{i0}} \right)\\
		{A_i} =& {k_i}\left( {{l_{i\rm{l}}} - {l_{i\rm{l}0}}} \right) + {b_i}{\rm{d}}{l_{i\rm{l}}}\\
		{\rm{d}}{l_{i\rm{l}}} =& {\rm{d}}{{\bf{t}}_i} \cdot {{\bf{t}}_{i0}}\\
		{{\bf{t}}_i} =& {{\bf{p}}_{{\rm{l}}}} - {{\bf{p}}_i}\\
		{{\bf{t}}_{i0}} =& \frac{{{{\bf{t}}_i}}}{{\left\| {{{\bf{t}}_i}} \right\|}}\\
		{l_{i\rm{l}}} =& \left\| {{{\bf{t}}_i}} \right\|,
	\end{aligned} \right. 
\end{equation} 
where ${A_i}$ is the tension that arises from the subtle deformation of the cable, 
${f_{\max ,i}}$ represents the maximum tension the cable can withstand without fracturing,
${{\bf{t}}_{i0}}$ is the unit vector pointing from the load to the $i$th UAV,
${{\bf{t}}_{i}}$ is the vector from the load to the $i$th UAV,
${k_i}$ and ${b_i}$ are parameters associated with the material cable ties of the cable,
${l_{i\rm{l}}}$ and ${l_{i\rm{l}0}}$ denote the current length and the initial length of the cable.
The model (\ref{eq:1-3}) delineates the cable as devoid of mass, solely furnishing tension without imparting repulsion, 
capable of engendering minute deformations and featuring a maximum tension threshold \cite{QuanSR}.

\subsection{Virtual tube model for the UAV-cable-load system}

\begin{figure}[!h]
	\centerline{\includegraphics[width=0.4\textwidth]{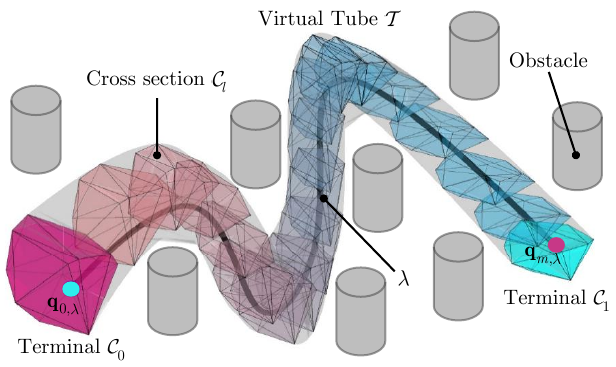}}
	\caption{The example of a virtual tube}
	\label{fig:tubemodel}
\end{figure}

\textit{Definitions of the Virtual Tube} \cite{mao2024optimal}:
a Virtual Tube $\mathcal{T}$ is defined as a set in $n$-dimension space represented by a 4-tuple $(\mathcal{C}_0,\mathcal{C}_1,\bf{f},\bf{h})$,
as shown in Fig. \ref{fig:tubemodel},
where 

\begin{itemize}
	\item $\mathcal{C}_0$ and $\mathcal{C}_1$ are disjoint bounded convex subsets in the $n$-dimension space, which called terminals.
	\item $\bf{f}$ is a diffeomorphism: $\mathcal{C}_0 \rightarrow \mathcal{C}_1$, 
	therefore, $\mathcal{P} = \left\{ ({\bf{q}}_0,{\bf{q}}_m)\vert {\bf{q}}_0 \in \mathcal{C}_0, {\bf{q}}_m =\bf{f}({\bf{q}}_0) \in \mathcal{C}_1  \right\}  $
	is a set of order pairs.
	\item $\bf{h}$ denotes the smooth map $\mathcal{P}\times \mathcal{L} \rightarrow \mathcal{T} $ where $\mathcal{L} = [0,1] $.
	${\bf h} \left( \left({\bf q}_0,{\bf q}_m\right),l\right)$ is called a trajectory for $({\bf{q}}_0,{\bf{q}}_m)$,
	\item The virtual tube $\mathcal{T}$ is defined as
	\begin{equation}\label{eq:1-5}
		{\mathcal T} = \{ {\bf h} \left( \left({\bf q}_0,{\bf q}_m\right),l\right) | \left({\bf q}_0,{\bf q}_m\right) \in {\mathcal P} , l \in {\mathcal L} \} .
	\end{equation} 
\end{itemize}

Moreover, there are some vital concepts of a virtual tube, such as the cross-section and the boundary of the virtual tube.
A cross-section $\mathcal{C}_l$ of $ {\mathcal T} $ at $l \in {\mathcal L} $ is expressed as 
$ {\mathcal{C}}_{l} = \left\{ {{\bf h}\left( {\left( {\bf q}_0,{\bf q}_m \right),{l}} \right)|\left({\bf q}_0,{\bf q}_m\right) \in {\mathcal{P}} } \right\} $.
The boundary of  $ {\mathcal T} $ is defined as $ \partial{\mathcal T}$ which represents the surface of the virtual tube \cite{mao2024optimal}.
The generator curve $\lambda(l)$ starting at ${\bf q}_{0,\lambda}$ and ending at ${\bf q}_{0,\lambda}$ is also denoted as ${\bf h} \left( \left({\bf q}_{0,\lambda},{\bf q}_{m,\lambda }\right),l\right)$, which is a speical trajectory.
For an object at any position ${\bf{p}}_i$ in the virtual tube ${\mathcal T}$, there is a closest point ${\bf{p}}_i^* \in \lambda$ on $\lambda$, 
that satisfies

\begin{equation}\label{eq:1-6}
	{\bf{p}}_i^*(l) = \arg {\rm{ }}\mathop {\min }\limits_{{{\bf{p}}_\lambda } \in \lambda } \left\| {{\bf{p}}_i}-{{\bf{p}}_\lambda } \right\|^2.
\end{equation} 
The tangent vector of ${\bf{p}}_i^*(l)$ on the generator curve $\lambda$ is defined as

\begin{equation}\label{eq:1-7}
	{\bf{t}}_{\rm{c}}(l) = \nabla _ l {\bf{p}}_i^*(l)
\end{equation} 
where $\nabla _ l = {\rm{d}/d}l$ is a spatial differentiator \cite{lv2024mean}.

\subsection{Objective}
For the UAV-cable-load system(\ref{eq:1-1}-\ref{eq:1-4}) described above, design a distributed controller reliant on the propulsive force ${{\bf{T}}_i},i = 1,2,...,n $ as the control commands,
which could orchestrate the entirety of the system(\ref{eq:1-1}-\ref{eq:1-4}) to avoid collisions between UAVs and allocate loads judiciously,
while getting through the pre-planned virtual tube (\ref{eq:1-5}) which starts at $\mathcal{C}\left( {{\mathbf{p}}_{\text{0}}} \right)$ and ends at $\mathcal{C}\left( {{\mathbf{p}}_{\text{1}}} \right)$.

\section{Controller Design}\label{chp:controllerdesign}

\begin{figure*}[!htbp]
	\centerline{\includegraphics[width=0.7\textwidth]{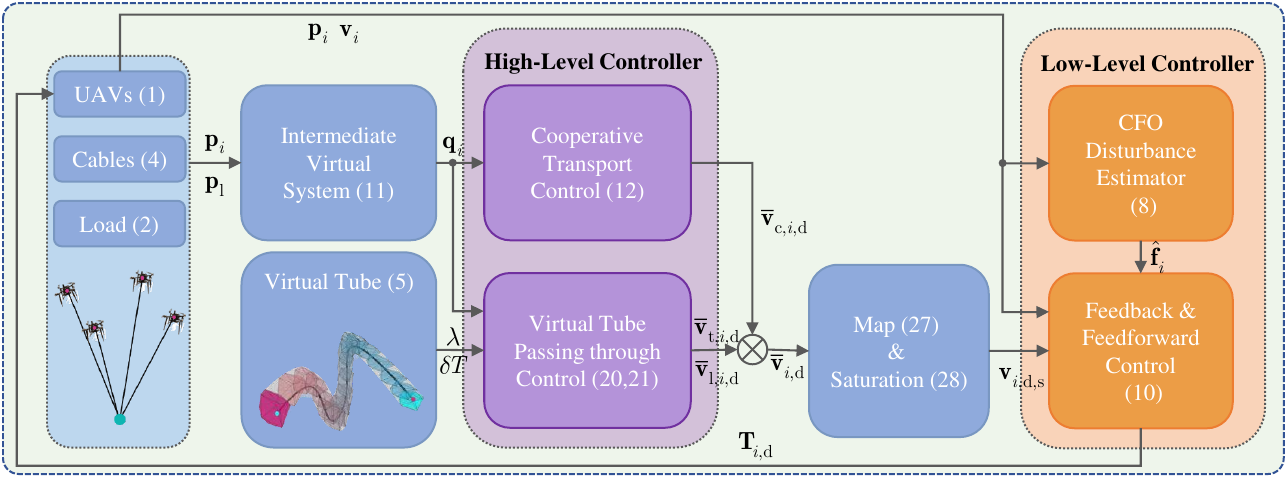}}
	\caption{Control diagram}
	\label{fig:Control diagram}
\end{figure*}

The control structure designed in this paper is illustrated in the Fig.\ref{fig:Control diagram}. Using a preplanned virtual tube(II.B) that avoids collisions with obstacles, a high-level controller provides velocity commands. This high-level controller consists of both the system transportation control(III.A) and the virtual tube passing through control(III.B). Furthermore, the velocity commands generated by the high-level controller are mapped and saturated before being input into the low-level controller(III.D). The low-level controller, according to the desired velocity and the UAVs' state, estimates the cable tension and performs feedforward and feedback control, regulating the UAVs' power system to output the desired tension.

\begin{figure}[!htbp]
	\centerline{\includegraphics[width=0.35\textwidth]{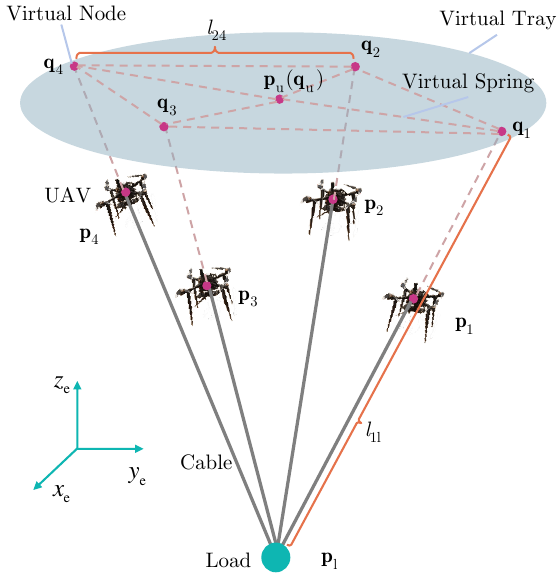}}
	\caption{An example of an intermediate system for the UAV-cable-load system}
	\label{fig:virtual sys}
\end{figure}

\subsection{Cooperative transport control}

Establish an intermediate system for the UAV-cable-load system, incorporating the \textit{virtual tray}, \textit{virtual nodes}, and \textit{virtual springs}, as shown in Fig. \ref{fig:virtual sys}.
The center of the virtual tray's coordinates is ${{\bf{q}}_{\rm{u}}} = [x_{{\rm{u}}} \ y_{{\rm{u}}} \ z_{{\rm{u}}}]^{\rm{T}} \in \mathbb{R}^3 $, the virtual node corresponding to the load is ${{\bf{q}}_{\rm{l}}} = [x_{{\rm{l}}} \ y_{{\rm{l}}} \ z_{{\rm{l}}}]^{\rm{T}} \in \mathbb{R}^3 $.
The coordinates of $i$th UAV's corresponding virtual node is denoted by ${{\mathbf{q}}_{i}}$ , satisfying the following relationships \cite{QuanSR}:
\begin{equation}\label{eq:2-4}
	{{\bf{q}}_i} = {{\bf{p}}_{\rm{l}}} + \frac{{\left| {{z_{\rm{u}}} - {z_{\rm{l}}}} \right|}}{{\left| {{z_{\rm{u}}} - {z_i}} \right|}}({{\bf{p}}_i} - {{\bf{p}}_{\rm{l}}}),i=1,2,...,n .
\end{equation}

In this intermediate system, the virtual tray is a sufficiently large but finite disk. 
The virtual nodes corresponding to the UAVs are all positioned on this virtual tray, 
meaning their heights are identical to the center height of the virtual tray, 
yet they can slide on the virtual tray. What's more, friction exists between the virtual nodes and the virtual tray.
All virtual nodes are interconnected by virtual springs, as well as the center of the virtual tray and the virtual nodes. 
Consequently, the desired velocity for the $i$th$(i= 1,2,\ldots,n,{\rm l},{\rm u})$ virtual node caused by the virtual spring between $i$th virtual node and $j$th$(j= 1,2,\ldots,n,{\rm l},{\rm u})$ virtual node is
\begin{equation}\label{eq:2-5}
	\begin{aligned}
		{{{\overline{\bf{v}}_{ij{,{\rm{d}}}}}}} &= - 
		{k_2} \begin{array}{l}
			{{w_{ij}}\left( {1 - \frac{{{\overline{l}_{ij0}}}}{{{\overline{l}_{ij}}}}} \right)({{\bf{q}}_i} - {{\bf{q}}_j})}
		\end{array} \\
		i,j &= {\rm{1,2,}}...{\rm{,}}n,{\rm l},{\rm u},i\neq j
	\end{aligned} 
\end{equation}
where ${k_i}>0$ is the elastic coefficient of the virtual spring,
${\bf{W}}=[w_{ij}]_{(n+2) \times (n+2)}$ is the connection matrix, while $w_{ij} = 1$ indicates that the $i$th virtual node and the $j$th one are connected.
A typical ${\bf{W}}$ is 
\begin{equation*}
	{\bf{W}}={\left[ {\begin{array}{*{20}{c}}
				0&1&1& \cdots &1&1\\
				1&0&1& \cdots &1&1\\
				1&1&0& \cdots &1&1\\
				\vdots & \vdots & \vdots & \ddots & \vdots & \vdots \\
				1&1&1& \cdots &0&0\\
				1&1&1& \cdots &0&0
		\end{array}} \right]_{(n + 2) \times (n + 2)}}.
\end{equation*}
${\overline{l}_{ij}}$ is the current length of the virtual spring,
${\overline{l}_{ij0}}$ is the initial length of the virtual spring.
For example, ${\overline{l}_{ij}}$ is defined as
\begin{equation}\label{eq:2-6}
	{{\overline{l}}_{ij}} = \sqrt {{{\left\| {{{\bf{q}}_i} - {{\bf{q}}_j}} \right\|}^2}} ,i,j = {\rm{1,2,}}...{\rm{,}}n,i\neq j .
\end{equation}
Meanwhile, ${\overline{l}_{ij0}}$ can be set according to the desired formation of the UAV-cable-load system 
or loaded based on the pre-takeoff position of the UAVs ${{\bf{q}}_{i0}}$ and other geometric constraints like lengths of cabels as
\begin{equation}\label{eq:2-7}
	{{\overline{l}}_{ij0}} = \sqrt {{{\left\| {{{\bf{q}}_{i0}} - {{\bf{q}}_{j0}}} \right\|}^2}} ,i,j = {\rm{1,2,}}...{\rm{,}}n,i\neq j .
\end{equation}

\textbf{Proposition 1.}  The intermediate system(\ref{eq:1-1}-\ref{eq:1-4},\ref{eq:2-4}) is stable if ${{{\overline{\bf{v}}_{ij}}}} = {{{\overline{\bf{v}}_{ij{,{\rm{d}}}}}}}$,
and one of the balance state satisfies $ {\overline{l}}_{ij} = {{\overline{l}}_{ij0}} $. 

\textbf{Proof.}
For any two virtual nodes $i,j = {\rm{1,2,}}...{\rm{,}}n,i\neq j$ in the intermediate system, 
a Lyapunov function is defined as 

\begin{equation}\label{eq:2-8}
	V_{{\rm{c}},ij} = \frac{1}{2}{\left( {{{\overline{l}}_{ij}} - {{\overline{l}}_{ij0}}} \right)^2} \ge 0 .
\end{equation}
And the derivative of the Lyapunov function is 

\begin{equation}\label{eq:2-9}
	\dot V_{{\rm{c}},ij} =  \left( {{{\overline{l}}_{ij}} - {{\overline{l}}_{ij0}}} \right){\dot {\overline{l}}_{ij}} .
\end{equation}
Define the position vector between virtual nodes $i,j$ as ${{\bf{q}}_{ij}} = {{\bf{q}}_i} - {{\bf{q}}_j}$, 
yield  ${\overline{l}}_{ij} = \sqrt {{\bf{q}}_{ij}^{\rm{T}}{{\bf{q}}_{ij}}} $.
Therefore, $\dot V_{{\rm{c}},ij}$ is expressed as

\begin{equation}\label{eq:2-10}
	\dot V_{{\rm{c}},ij}  = \left( {{{\overline{l}}_{ij}} - {{\overline{l}}_{ij0}}} \right)\frac{{{\bf{q}}_{ij}^{\rm{T}}{{{\bf{\dot q}}}_{ij}}}}{{\sqrt {{\bf{q}}_{ij}^{\rm{T}}{{\bf{q}}_{ij}}} }}.
\end{equation}
Due to
\begin{equation}\label{eq:2-11}
	\begin{aligned}
		{{{\bf{\dot q}}}_{ij}} &= {{{\bf{\dot q}}}_i} - {{{\bf{\dot q}}}_j} = {\overline {\bf{v}}_{ij,{\rm{d}}}} - {\overline {\bf{v}}_{ji,{\rm{d}}}}\\
		&= -2k w_{ij} \left( {1 - \frac{{{{\overline{l}}_{ij0}}}}{{{{\overline{l}}_{ij}}}}} \right){{\bf{q}}_{ij}}
	\end{aligned}
\end{equation}
Equ. (\ref{eq:2-10}) is reduced to
\begin{equation}\label{eq:2-12}
	\begin{aligned}
		\dot V_{{\rm{c}},ij} &= \left( {{{\overline{l}}_{ij}} - {{\overline{l}}_{ij0}}} \right)\frac{{{\bf{q}}_{ij}^{\rm{T}}}}{{\sqrt {{\bf{q}}_{ij}^{\rm{T}}{{\bf{q}}_{ij}}} }}(-2k w_{ij})\left( {1 - \frac{{{{\overline{l}}_{ij0}}}}{{{{\overline{l}}_{ij}}}}} \right){{\bf{q}}_{ij}}\\
		&= -2k w_{ij}\frac{{{{\left( {{{\overline{l}}_{ij}} - {{\overline{l}}_{ij0}}} \right)}^2}}}{{{{\overline{l}}_{ij}}}}\frac{{{\bf{q}}_{ij}^{\rm{T}}{{\bf{q}}_{ij}}}}{{\sqrt {{\bf{q}}_{ij}^{\rm{T}}{{\bf{q}}_{ij}}} }} \leq 0. 
	\end{aligned}
\end{equation}
Meanwhile, if $ {\overline{l}}_{ij} = {{\overline{l}}_{ij0}} $, then $V_{{\rm{c}},ij}=0$ and $\dot V_{{\rm{c}},ij}=0$. $\square$

\textbf{Remark 1.}  The proposition and proof above reveal that this virtual intermediate system has a trend 
of converging on a tendency to restore the individual virtual springs to their initial state and to remain in that state.
This is pivotal for UAV cooperative transportation. This means that the researcher can reasonably adjust the initial length of the virtual spring of the intermediate virtual system by taking into account the physical constraints, such as the length of the cables and the load capacity of the UAVs so that the UAV cooperative transportation system better distributes the pulling force provided by each UAV. During the process of transportation, even if there is interference from external forces, the system still tends to recover to the state of the initial setup and, therefore, can better carry out the cooperative transportation tasks.

\subsection{Virtual tube passing through control}

In this paper, we limit our consideration to the case where the virtual tube is two-dimensional \cite{quan2021practical},
i.e., ${\mathbf{{q}}}_{0},{\mathbf{{q}}}_{m} \in \mathbb{R}^2$. 
In this case, the virtual tube's boundary $ \partial \mathcal{T} $ is a couple of curves.
The virtual tube is preplanned to ensure there are no obstacles in the tube.
To control the UAV-cable-load system get through the virtual tube from the start terminal $\mathcal{C}_0$ 
to the end terminal $\mathcal{C}_1$ for completing the transport mission, 
the following control component needs to be incorporated into the velocity commands of the virtual nodes to guide the UAV-cable-load system to the end terminal:
\begin{equation}\label{eq:2-13}
	{\overline{\mathbf{v}}_{\text{l},i,\text{d}}}={{\rm{sat}}\left( {{k_1}l\left( {{{{\bf{\bar q}}}_i}} \right)\eta \left( {{{{\bf{\bar q}}}_i}} \right){{\bf{t}}_{\rm{c}}}\left( {{{{\bf{\bar q}}}_i}} \right),{v_{m,i}}} \right)}
\end{equation}
where ${ {\mathbf{\overline{q}}}_{i} }$ denotes the coordinates of the $i$th virtual nodes in the $oxy$ plane, 
and ${ {\mathbf{t}}_{\text{c}}}\left( {{\mathbf{\overline{q}}}_{i}} \right)$ represents the unit tangent vector at ${ {\mathbf{\overline{q}}}_{i} }$ pointing in the direction of the generating curve,
${ {v}_{\text{m},\text{l}} }$ is the magnitude control component steering the UAV-cable-load system towards end terminal $\mathcal{C}_0$, 
which can be adjusted based on the system's operational efficiency.
The corresponding line integral Lyapunov function for approaching the end terminal is designed as follows:

\begin{equation}\label{eq:2-13-1}
	{V_{{\rm{l}},i}} = \int_0^t {{\rm{sat}}{{\left( {{k_1}l\left( {{{\bf{\overline{q}}}_i}} \right)\eta \left( {{{\bf{\overline{q}}}_i}} \right){{\bf{t}}_{\rm{c}}}\left( {{{\bf{\overline{q}}}_i}} \right),{v_{m,i}}} \right)}^T}{\overline{\mathbf{v}}_{i,\text{d}}}{\rm{d}}\tau }
\end{equation}
where $l\left( {{{\bf{\overline{q}}}_i}} \right),\eta \left( {{{\bf{\overline{q}}}_i}} \right)$ have been previously defined and explained in our earlier work \cite{quan2023distributed}.

In order to ensure the UAV-cable-load system within the virtual tube and prevent collisions with obstacles outside the virtual tube,
it is necessary to incorporate the following control components into the velocity command of the virtual nodes to prevent the UAV-cable-load system from getting out of the virtual tube:
\begin{equation}\label{eq:2-14}
	{\overline{\mathbf{v}}_{\text{t},i,\text{d}}}=-\left( {{\mathbf{I}}_{2}}-{{\mathbf{t}}_{\text{c}}}\left( {{\mathbf{\overline{q}}}_{i}} \right)\mathbf{t}_{\text{c}}^{\text{T}}\left( {{\mathbf{\overline{q}}}_{i}} \right) \right){{\mathbf{c}}_{i}}.
\end{equation}
Here   
\begin{equation}\label{eq:2-15}
	{{\mathbf{c}}_{i}}=\frac{\partial {{V}_{\text{t},i}}}{\partial {{d}_{\text{t,}i}}}{{\left( \frac{\partial {{r}_{\text{t}}}\left( {{\mathbf{\overline{q}}}_{i}} \right)}{\partial {{\mathbf{\overline{q}}}_{i}}}-\frac{{{\left( {{\mathbf{\overline{q}}}_{i}}-\mathbf{m}\left( {{\mathbf{\overline{q}}}_{i}} \right) \right)}^{\text{T}}}}{\left\| {{\mathbf{\overline{q}}}_{i}}-\mathbf{m}\left( {{\mathbf{\overline{q}}}_{i}} \right) \right\|}\left( {{\mathbf{I}}_{\text{3}}}-\frac{\partial \mathbf{m}\left( {{\mathbf{\overline{q}}}_{i}} \right)}{\partial {{\mathbf{\overline{q}}}_{i}}} \right) \right)}^{\text{T}}}
\end{equation}
where ${{\mathbf{\overline{q}}}_{i}} = [x_i \ y_i ]^{\rm{T}}$ denotes the coordinates of the $i$th virtual node in the $oxy$ plane,
${{\mathbf{t}}_{\text{c}}}\left( {{\mathbf{\overline{q}}}_{i}}\right)$ represents the unit tangent vector at \({{\mathbf{\overline{q}}}_{i}}\) pointing in the direction of the generating line, 
\({{r}_{\text{t}}}\left( {{\mathbf{\overline{q}}}_{i}} \right)\) signifies one-half of the virtual tube's width at \({{\mathbf{\overline{q}}}_{i}}\),
\(\mathbf{m}\left( {{\mathbf{\overline{q}}}_{i}} \right)\) indicates the midpoint of the cross-section \(\mathcal{C}\left( {{\mathbf{\overline{q}}}_{i}} \right)\) at \({{\mathbf{\overline{q}}}_{i}}\), 
and \(\partial {{V}_{\text{t},i}}\) delineates the corresponding Lyapunov-like barrier function of the virtual tube at \({{\mathbf{\overline{q}}}_{i}}\):

\begin{equation}\label{eq:2-15-1}
	{V_{{\rm{t}},i}} = \frac{{{k_3}{\sigma _{\rm{t}}}\left( {{d_{{\rm{t}},i}}} \right)}}{{\left( {1 + {\varepsilon _{\rm{t}}}} \right){d_{{\rm{t}},i}} - {r_{\rm{s}}}s\left( {\frac{{{d_{{\rm{t}},i}}}}{{{r_{\rm{s}}}}},{\varepsilon _{\rm{s}}}} \right)}}
\end{equation}
where $\sigma _{\rm{t}},{d_{{\rm{t}},i}},\varepsilon _{\rm{t}},{r_{\rm{s}}},s$ have been previously defined and explained in our earlier work \cite{quan2023distributed}.

\subsection{Synthesis of control components and convergence analysis}

The cooperative transport controller for passing narrow areas based on the virtual tube for $i$th virtual nodes is represented as
\begin{equation}\label{eq:2-16}
	\begin{aligned}
		{\overline{\bf{v}}_{i,{\rm{d}}}} &=-\left( {{\overline{\bf{v}}_{{\rm{l}},i,{\rm{d}}}}} - {{\overline{\bf{v}}_{{\rm{c,}}i,{\rm{d}}}}} + {{\overline{\bf{v}}_{{\rm{t}},i,\text{d}}}}\right) \\
		&=-\left(  {\rm{sat}}\left( {{k_1}l\left( {{{{\bf{\bar q}}}_i}} \right)\eta \left( {{{{\bf{\bar q}}}_i}} \right){{\bf{t}}_{\rm{c}}}\left( {{{{\bf{\bar q}}}_i}} \right),{v_{m,i}}} \right) \right. \\
		&\quad+ {k_2}\sum\limits_{j \in {{\mathcal{N} }_{m,i}}} {{w_{ij}}\left( {1 - \frac{{{{\bar l}_{ij0}}}}{{{{\bar l}_{ij}}}}} \right)({{{\bf{\bar q}}}_i} - {{{\bf{\bar q}}}_j})} \\
		& \left. \quad- \left( {{{\bf{I}}_2} - {{\bf{t}}_{\rm{c}}}\left( {{{{\bf{\bar q}}}_i}} \right){\bf{t}}_{\rm{c}}^{\rm{T}}\left( {{{{\bf{\bar q}}}_i}} \right)} \right){{\bf{c}}_i} \right)  
	\end{aligned}
\end{equation}
where ${{\mathcal{N} }_{m,i}} = \left\{ j = {\rm{1,2,}}...{\rm{,}}n,\rm{u},\rm{l},\ j\neq i \right\} $ is defined as a set of all marks of other virtual nodes for the $i$th virtual node.

\textbf{Proposition 2.} If ${{{\overline{\bf{v}}_{i}}}} = {{{\overline{\bf{v}}_{i{,{\rm{d}}}}}}}$ and the virtual tube $\mathcal{T}$ is wide enough to accommodate at least one UAV, the intermediate system(\ref{eq:1-1}-\ref{eq:1-4},\ref{eq:2-4}) is stable and can get through $\mathcal{T}$ which starts at $\mathcal{C}\left( {{\mathbf{p}}_{\text{0}}} \right)$ and ends at $\mathcal{C}\left( {{\mathbf{p}}_{\text{1}}} \right)$ without collision between the virtual node and the boundary of $\mathcal{T}$.

\textbf{Proof.}

A Lyapunov-like function is defined as
\begin{equation}\label{eq:2-17}
	V = \sum\limits_{i = 1}^M ({ {{V_{{\rm{l}},i}} + \frac{1}{2} k_2 \sum\limits_{j = 1,j \ne i}^M {w_{i,j}{V_{{\rm{c}},ij}}}  + {V_{{\rm{t}},i}}}} ) \geq 0
\end{equation}
where $M=n+2$ represents the number of elements in set ${{\mathcal{N} }_{m,i}}$. According to our previous work in \cite{quan2023distributed} and Equ.(\ref{eq:2-10}), $\dot V$ is shown as

\begin{equation}\label{eq:2-18}
	\begin{aligned}
		\dot V &= \sum\limits_{i = 1}^M {} ({\rm{sat}}{\left( {{k_1}l\left( {{{{\bf{\bar q}}}_i}} \right)\eta \left( {{{{\bf{\bar q}}}_i}} \right){{\bf{t}}_{\rm{c}}}\left( {{{{\bf{\bar q}}}_i}} \right),{v_{m,i}}} \right)^{\rm{T}}}{{{\bf{\bar v}}}_{i,{\rm{d}}}} \\
		&\quad+  \frac{1}{2}{k_2}\sum\limits_{j = 1,j \ne i}^M {{w_{ij}}\left( {{{\bar l}_{ij}} - {{\bar l}_{ij0}}} \right)\frac{{{\bf{\bar q}}_{ij}^{\rm{T}}{{{\bf{\dot{\bar{q}}}}}_{ij}}}}{{\sqrt {{\bf{\bar q}}_{ij}^{\rm{T}}{{{\bf{\bar q}}}_{ij}}} }}} \\ 
		&\quad-  {\left( {\left( {{{\bf{I}}_2} - {{\bf{t}}_{\rm{c}}}\left( {{{{\bf{\bar q}}}_i}} \right){\bf{t}}_{\rm{c}}^{\rm{T}}\left( {{{{\bf{\bar q}}}_i}} \right)} \right){{\bf{c}}_i}} \right)^{\rm{T}}}{{{\bf{\bar v}}}_{i,{\rm{d}}}}) \\
		& = \sum\limits_{i = 1}^M {} ({\rm{sat}}{\left( {{k_1}l\left( {{{{\bf{\bar q}}}_i}} \right)\eta \left( {{{{\bf{\bar q}}}_i}} \right){{\bf{t}}_{\rm{c}}}\left( {{{{\bf{\bar q}}}_i}} \right),{v_{m,i}}} \right)^{\rm{T}}}{{{\bf{\bar v}}}_{i,{\rm{d}}}} \\
		&\quad+ \frac{1}{2}{k_2}\sum\limits_{j = 1,j \ne i}^M {{w_{ij}}\left( {{l_{ij}} - {l_{ij0}}} \right)\frac{{{\bf{\bar q}}_{ij}^{\rm{T}}}}{{\sqrt {{\bf{\bar q}}_{ij}^{\rm{T}}{{{\bf{\bar q}}}_{ij}}} }}} \left( {{{{\bf{\bar v}}}_{i,{\rm{d}}}} - {{{\bf{\bar v}}}_{j,{\rm{d}}}}} \right) \\
		&\quad- {\left( {\left( {{{\bf{I}}_2} - {{\bf{t}}_{\rm{c}}}\left( {{{{\bf{\bar q}}}_i}} \right){\bf{t}}_{\rm{c}}^{\rm{T}}\left( {{{{\bf{\bar q}}}_i}} \right)} \right){{\bf{c}}_i}} \right)^{\rm{T}}}{{{\bf{\bar v}}}_{i,{\rm{d}}}}) \\
		&=\sum\limits_{i = 1}^M {} ({\rm{sat}}{\left( {{k_1}l\left( {{{{\bf{\bar q}}}_i}} \right)\eta \left( {{{{\bf{\bar q}}}_i}} \right){{\bf{t}}_{\rm{c}}}\left( {{{{\bf{\bar q}}}_i}} \right),{v_{m,i}}} \right)^{}} \\
		&\quad+ {k_2}\sum\limits_{j \in {{\rm N}_{m,i}}}^{} {{w_{ij}}\left( {{l_{ij}} - {l_{ij0}}} \right)\frac{{{\bf{\bar q}}_{ij}^{}}}{{\sqrt {{\bf{\bar q}}_{ij}^{\rm{T}}{{{\bf{\bar q}}}_{ij}}} }}}  \\
		&\quad- { {\left( {{{\bf{I}}_2} - {{\bf{t}}_{\rm{c}}}\left( {{{{\bf{\bar q}}}_i}} \right){\bf{t}}_{\rm{c}}^{\rm{T}}\left( {{{{\bf{\bar q}}}_i}} \right)} \right){{\bf{c}}_i}} })^{\rm{T}}{{{\bf{\bar v}}}_{i,{\rm{d}}}} .
	\end{aligned} 
\end{equation}
Due to ${\overline{l}}_{ij} = \sqrt {{\bf{q}}_{ij}^{\rm{T}}{{\bf{q}}_{ij}}} $, one has

\begin{equation}\label{eq:2-18-1}
	\begin{aligned}
		\dot V &= \sum\limits_{i = 1}^M {} ({\rm{sat}}{\left( {{k_1}l\left( {{{{\bf{\bar q}}}_i}} \right)\eta \left( {{{{\bf{\bar q}}}_i}} \right){{\bf{t}}_{\rm{c}}}\left( {{{{\bf{\bar q}}}_i}} \right),{v_{m,i}}} \right)^{}} \\
		&\quad+ {k_2}\sum\limits_{j \in {{\rm N}_{m,i}}}^{} {{w_{ij}}\left( {1 - \frac{{{l_{ij0}}}}{{{l_{ij}}}}} \right){\bf{\bar q}}_{ij}^{}} \\
		&\quad- { {\left( {{{\bf{I}}_2} - {{\bf{t}}_{\rm{c}}}\left( {{{{\bf{\bar q}}}_i}} \right){\bf{t}}_{\rm{c}}^{\rm{T}}\left( {{{{\bf{\bar q}}}_i}} \right)} \right){{\bf{c}}_i}} })^{\rm{T}}{{{\bf{\bar v}}}_{i,{\rm{d}}}} .
	\end{aligned} 
\end{equation}
By applying the desired velocity command(\ref{eq:2-16}) to all virtual nodes, $\dot V$ satisfies $\dot V \leq 0$, which implies that the controller is stable while simultaneously fulfilling the functionalities of passing through and collision avoidance mentioned above.
The proof that the Lyapunov-like function of the form given in Equ.(\ref{eq:2-16}) ensures the swarm passing through the virtual tube without colliding with the boundary is similar in \textit{Theorem 1} of our previous work \cite{quan2023distributed} and will be omitted here. $\square$

\renewcommand{\floatpagefraction}{.95} 
\renewcommand{\textfraction}{.05} 

\subsection{Implement}

Based on the geometric relationship between the intermediate virtual system and the real UAV-cable-load system and 
assuming the altitudes of the virtual tray and load remain constant, 
the desired velocity of the virtual nodes can be mapped to the desired velocity of the UAVs as

\begin{equation}\label{eq:2-20}
	{{\bf{v}}_{i,{\rm d}}} = {{\bf{v}}_{\rm{l}}} + \frac{{\left| {{z_{\rm{u}}} - {z_i}} \right|}}{{\left| {{z_{\rm{u}}} - {z_{\rm{l}}}} \right|}}({\overline{\bf{v}}_{i,{\rm{d}}}} - {{\bf{v}}_{\rm{l}}}),i=1,2,...,n .
\end{equation}

In order to guarantee safety, the saturation with maximum velocity ${v_{{\rm{m}},i}}$ is set in the velocity controller:
\begin{equation}\label{eq:2-21}
	{{\bf{{v} }}_{i,{\rm{d,s}}}} = {\rm{sat}}\left( {{\bf{v}}_{i,{\rm{d}}}},{v_{{\rm{m}}}} \right)
\end{equation}
where the saturation function is defined as 
\begin{equation}\label{eq:2-22}
	{\rm{sat}}\left( {{{\bf{v}}_{i,{\rm{d}}}},{v_{{\rm{m}}}}} \right) = \left\{ {\begin{array}{*{20}{c}}
			{{{\bf{v}}_{i,{\rm{d}}}}}&{\left\| {{{\bf{v}}_{i,{\rm{d}}}}} \right\| \le {v_{{\rm{m}}}}}\\
			{{v_{{\rm{m}}}}\frac{{{{\bf{v}}_{i,{\rm{d}}}}}}{{\left\| {{{\bf{v}}_{i,{\rm{d}}}}} \right\|}}}&{\left\| {{{\bf{v}}_{i,{\rm{d}}}}} \right\| > {v_{{\rm{m}}}}} . 
	\end{array}} \right.
\end{equation}

Then, the objective of the low-level controller is: given a desired velocity ${\bf{v}}_{i,\rm{d,s}}$ 
design a proper ${\bf{T}}_{i,\rm{d}}$ for model (\ref{eq:1-1}) to ensure

\begin{equation}\label{eq:2-2}
	\lim_{t \to \infty}\left\| {{\bf e_v} (t)}\right\|=0, {\bf e_v} \triangleq {\bf{v}}_{i,\rm{d,s}}-{\bf{v}}_{i}.
\end{equation}

According to Equ.(\ref{eq:1-1}), ${\bf{T}}_i$ is the input, ${\bf{G}}_i$ can be obtained from known mass parameters.
Then, the problem becomes how to obtain ${\bf{f}}_i$. 
In practice, ${\bf{f}}_i$ is the main disturbance UAV subjected during the transportation process.
It can be measured using a force sensor or estimated through a disturbance observer like the extended state observer(ESO), compensation function observer(CFO) \cite{qi2021problems}, and others.
In this paper, CFO is used to estimate ${\bf{f}}_i$. 

Then, with the estimation of ${\bf{f}}_i$, denoted as $\hat{\bf{f}}_i$, the low-level controller contained feedback and feedforward terms can be designed as
\begin{equation}\label{eq:2-3}
	{\bf{T}}_{i,\rm{d}} = m_i ( {\bf{K}}_{{\bf{v}}{\rm p}}{\bf e_v} + {\bf{K}}_{{\bf{v}}{\rm i}}\int {\bf e_v} + {\bf{K}}_{{\bf{v}}{\rm d}}{\dot{{\bf e}}_{\bf {v}}} ) + \hat{\bf{f}}_i - {\bf G}_i
\end{equation}
where ${\bf{K}}_{{\bf{v}}{\rm p}},{\bf{K}}_{{\bf{v}}{\rm i}},{\bf{K}}_{{\bf{v}}{\rm d}} \in \mathbb{R} ^{3 \times 3}$ are diagonal positive definite matrices.
Assuming that $\dot{\bf{v}}_{i,\rm{d}} = {\bf{0}}_{3 \times 1}$, if $\lim_{t \to \infty}\left\lvert {{\bf e_{T}} (t)}\right\rvert=0$ where ${\bf e_T} \triangleq {\bf{T}}_{i,\rm{d}}-{\bf{T}}_{i}$, 
then $\lim_{t \to \infty}\left\lvert {{\bf e_{v}} (t)}\right\rvert=0$.

\textbf{Remark 2.}  The proposed controller is completely distributed. This controller requires only the positions of other UAVs and the load, along with its own state. Whether through communication or non-communicative visual sensors, it can operate autonomously.  
Furthermore, this controller does not require the cable lengths between the UAV and the load to remain identical and can tolerate uncertainties in the cable length.

\section{Numerical Simulation and Analysis}

\renewcommand{\floatpagefraction}{1} 
\renewcommand{\textfraction}{.0} 

\begin{figure*}[htbp]
	\centerline{\includegraphics[width=0.85\textwidth]{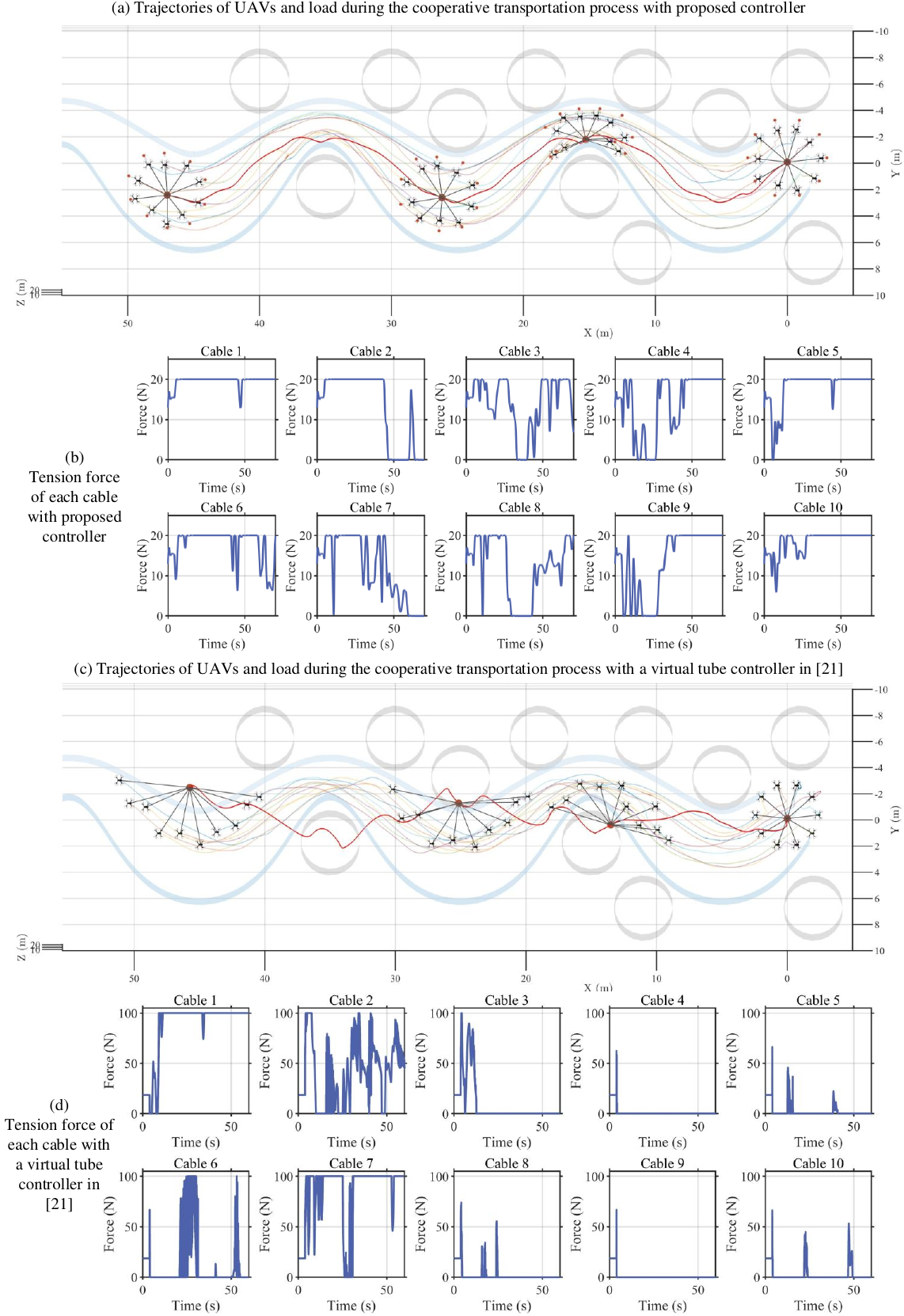}}
	\caption{Comparison between the proposed controller and the virtual tube controller \cite{quan2023distributed} in simulation}
	\label{fig:simall}
\end{figure*}

\subsection{Simulation Result}

This section presents numerical simulation results to validate the performance and robustness of the proposed control method. The simulations are conducted in a $50m\times20m\times20m$ complex environment featuring densely distributed obstacles and narrow passages between them. Based on the distribution of the obstacles, a virtual tube is generated with our previous work \cite{mao2022making}. This virtual tube ensures that as long as the UAV-cable-load system operates within it, collisions with obstacles are avoided. A video has been uploaded to the URL: \url{https://youtu.be/_VUCs72NQ9s} to introduce the paper briefly and show the simulation result visually.

\begin{table}[h!]
	\centering
	\caption{Environment parameters and system parameters}
	\begin{tabular}{lll}
		\hline
		\hline
		Parameter & Description & Values\\
		\hline
		$\mathfrak{L}_{\rm{env}} (\text{m}^3) $ & Environmental scale & $50*20*20$\\
		$n_{\rm{obs}} $ & Number of obstacles & $11$\\
		$m_i (\text{kg})$ & Mass of UAVs &  $2$\\
		$m_{\rm{l}} (\text{kg})$ & Mass of load & $ 18 $\\
		$f_{i,\text{max}} (\text{N})$ & Maximum load capacity & $20$\\
		$l_{i\rm{l}} (\text{m})$ & Cable length  & $8.3, 7.0, 8.6, 8.8, 8.3,$ \\ & & $ 8.5, 8.4, 7.7, 8.3, 7.3$ \\
		\hline
		\hline
	\end{tabular}
	\label{model_parameter_table}
\end{table}

Up to ten UAVs are utilized to cooperatively transport a load through this challenging environment. Each UAV has a mass of $ m_i = 2 \rm{kg} $ and a maximum load capacity of 20 N.  
The load has a mass of $ m_{\rm{l}} = 18 \rm{kg} $, which is significantly heavy compared to the maximum cumulative load capacity of 200 N for the ten UAVs. This highlights a key distinction of this work compared to previous studies, where much smaller load weights were considered compared to the load capacity of UAVs. The cable lengths are ser as Tab. \ref{model_parameter_table}, with each cable having a distinct length. This reflects that the control method proposed in this paper does not require uniform cable lengths. 
The control parameters in the simulation are set as Tab. \ref{control_parameter_table}. 

\begin{table}[h!]
	\centering
	\caption{Controller parameters}
	\renewcommand{\arraystretch}{1.2} 
	\setlength{\tabcolsep}{3pt}      
	\begin{tabular}{c|lll}			
		\hline
		\hline
		\multicolumn{1}{l|}{}                                                             & Paramters & Simulation & Experiment \\ 
		\hline
		& $ k_1 $ & 1.0 & 0.3 \\
		& $ k_2$ & 1.0 & 0.1 \\
		& $ k_3 $ & 1.0 & 1.0 \\
		\multirow{-4}{*}{\begin{tabular}[c]{@{}c@{}}High-Level \\ Controller \end{tabular}}         & $ v_{\rm{m}} $ & 1.5  & 1.0 \\ 
		\hline
		
		\rowcolor[HTML]{EFEFEF} 
		\rowcolor[HTML]{EFEFEF}
		\cellcolor[HTML]{EFEFEF}                                                          & $ {\bf{K}}_{{\bf{v}}{\rm p}} $ & $\left[ {\begin{array}{*{20}{c}}2&2&2\end{array}} \right]^{\rm{T}}$  & $\left[ {\begin{array}{*{20}{c}}2&2&2\end{array}} \right]^{\rm{T}}$ \\
		\rowcolor[HTML]{EFEFEF} 
		\cellcolor[HTML]{EFEFEF}                                                          & $ {\bf{K}}_{{\bf{v}}{\rm i}} $ & $\left[ {\begin{array}{*{20}{c}}0.1&0.1&0.1\end{array}} \right]^{\rm{T}}$  & $\left[ {\begin{array}{*{20}{c}}0.1&0.1&0.1\end{array}} \right]^{\rm{T}}$ \\
		\rowcolor[HTML]{EFEFEF} 
		\multirow{-4}{*}{\cellcolor[HTML]{EFEFEF}\begin{tabular}[c]{@{}c@{}}Low-Level \\ Controller\end{tabular}}      
		& $ {\bf{K}}_{{\bf{v}}{\rm d}} $ & $\left[ {\begin{array}{*{20}{c}}0.1&0.1&0.1\end{array}} \right]^{\rm{T}}$  & $\left[ {\begin{array}{*{20}{c}}0.1&0.1&0.1\end{array}} \right]^{\rm{T}}$  \\ 
		\hline
		\hline
	\end{tabular}
	\label{control_parameter_table}
\end{table}

\begin{figure}[h!]
	\centerline{\includegraphics[width=0.5\textwidth]{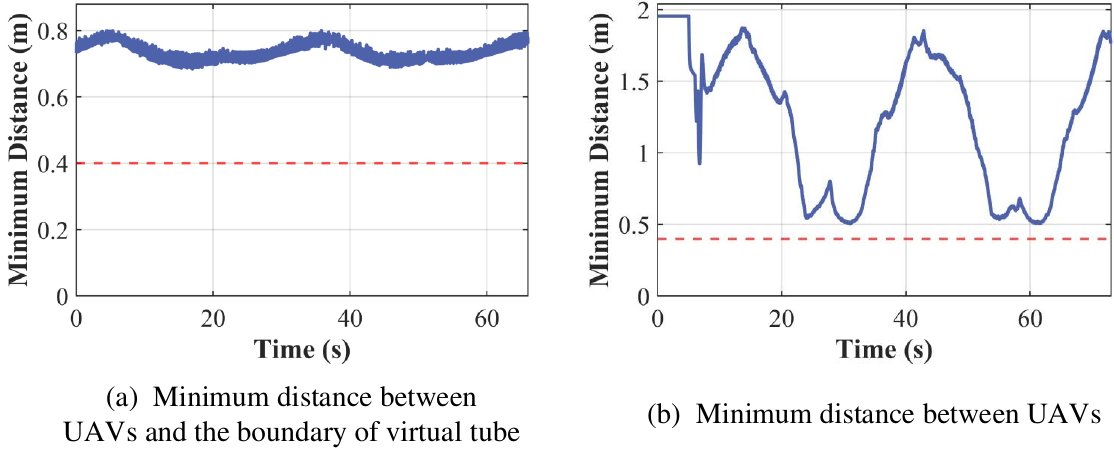}}
	\caption{Minimum distance for safety}
	\label{fig:simdis}
\end{figure}

The numerical simulation results are shown in Fig.\ref{fig:simall}.(a) and (b), demonstrating that the ten UAVs can cooperatively and smoothly transport the load through the complex environment. 
Fig.\ref{fig:simall}.(a) illustrates the trajectory of the entire system as it traverses the complex environment, along with the initial and final formations. A video link is provided here to demonstrate the performance of the proposed controller.  Initially, the virtual nodes corresponding to the UAVs form an approximate circle with a radius of 3 m. It is evident that when the cooperative transportation system passes through narrow sections of the virtual tube, the virtual intermediary system is compressed by the boundary of the virtual tube. As a result, the initially circular formation is flattened, prompting the UAVs to adjust themselves accordingly, thereby avoiding collisions with obstacles in these narrow regions.
When the system passes through wider sections of the virtual tube, the virtual nodes tend to return to their initial formation, and the UAVs adjust correspondingly, restoring the initial state where the cable tensions are more evenly distributed. Throughout the process, the cooperative transportation system operates stably and smoothly.

Fig.\ref{fig:simdis} shows the minimum distance between the UAVs and the virtual tube boundary and the minimum distance between the UAVs themselves during the cooperative transportation process through the complex environment. For safety reasons, to prevent collisions, the minimum distance mentioned above should not be less than 0.4 meters.
From Fig.\ref{fig:simdis}.(a), it can be observed that the minimum distance between the UAVs and the tube boundary consistently remains above 0.68 m. This indicates that the UAVs maintain a safe distance from the virtual tube boundary, ensuring that they do not collide with any obstacles. It should be noted that the curve in Fig.\ref{fig:simdis}.(a) appears jagged, which is due to the fact that the UAV closest to the tube boundary changes at different moments.
From Fig.\ref{fig:simdis}.(b), it can be observed that the minimum distance between the UAVs remains above 0.5 m throughout the process, ensuring that there is no collision between the UAVs.

\renewcommand{\floatpagefraction}{.95} 
\renewcommand{\textfraction}{.05} 
\begin{figure*}[h!]
	\centerline{\includegraphics[width=0.8\textwidth]{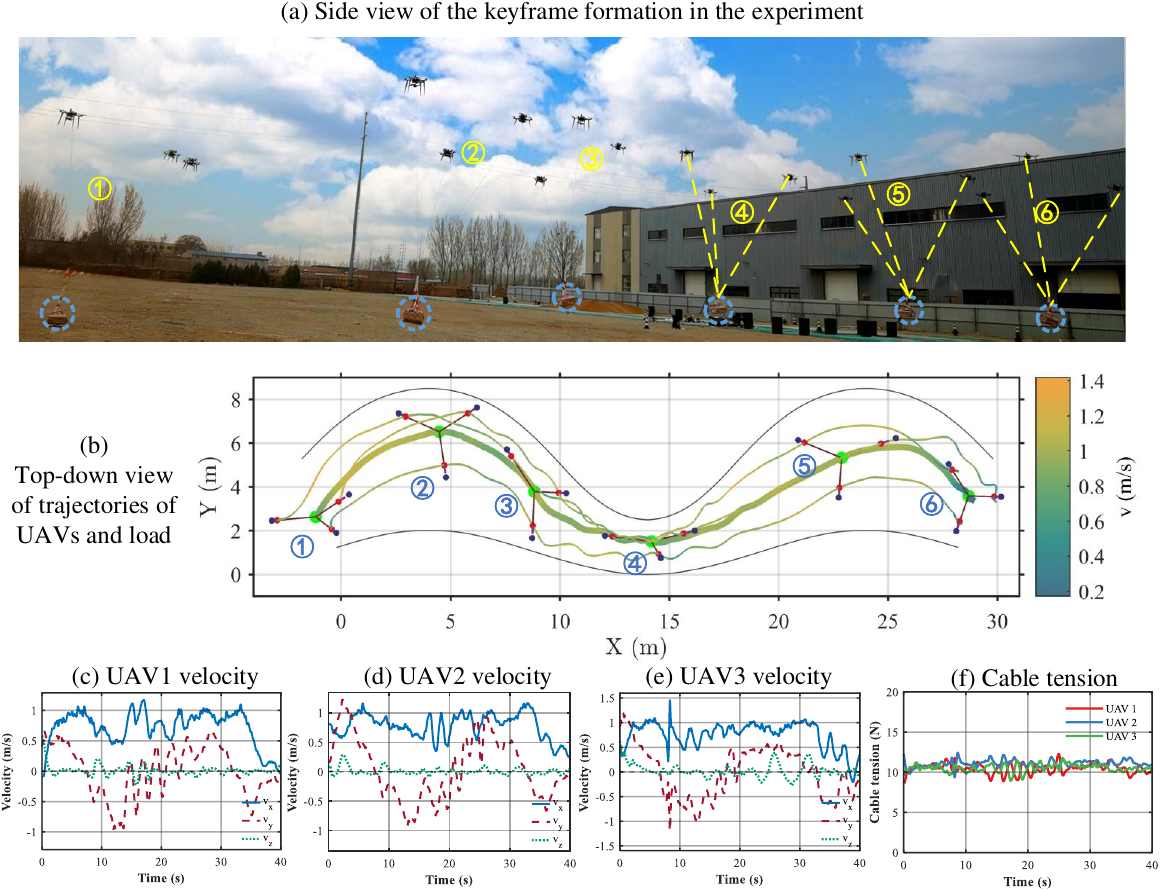}}
	\caption{Experiment results}
	\label{fig:expall}
\end{figure*}

Fig.\ref{fig:simall}.(b) illustrates the tension in each cable during the cooperative transportation process through the complex environment, which represents the force provided by each UAV on the load. From Fig.\ref{fig:simall}.(b), it can be observed that the cable tension is constrained to the maximum pulling force of 20 N that the UAVs can provide. 
During the transportation process, there are moments when the tension in some cables is relatively small. This is because, in sections where the virtual tube becomes too narrow, collision avoidance takes a higher priority. For a large-scale cooperative transportation scenario with 10 UAVs and significant over-constraint conditions, the variations in cable tension, as shown in the figure, are relatively smooth and reasonably distributed. This demonstrates that the proposed control method achieves effective results.

\subsection{Comparison with virtual tube control}

In our previous work \cite{QuanSR}, we compared the dissipative system-based multi-UAV cooperative transportation method with the formation-based or planning-based cooperative transportation methods. In this paper, the focus is primarily on comparing the proposed method with the traditional virtual tube control method to highlight the necessity and advantages of the theory presented.

As shown in Fig.\ref{fig:simall}.(c), when the traditional virtual tube control method is used, it causes instability in the UAV-cable-load system. Specifically, the UAV formation is squeezed, becoming elongated and dispersed. Even after passing through narrow areas into wider zones, the formation does not recover to one that facilitates cooperative transportation. Additionally, due to the unreasonable formation and the effect of the cables, the load's trajectory undergoes severe oscillations, potentially leading to the load exiting the virtual tube and colliding with obstacles, which creates significant safety risks. 

Furthermore, Fig.\ref{fig:simall}.(d) illustrates that with the traditional virtual tube control method alone, the forces by each UAV are not distributed reasonably. As a result, some UAVs end up bearing up to 100 N of tension, while others contribute almost no force throughout the transport process, and the cable tension fluctuates drastically.  It should be noted that, based on simulation trials, if the UAV's load capacity is limited to 20N, the virtual tube controller is unable to accomplish the transport task.

In conclusion, the traditional virtual tube control method is not well-suited for multi-UAV cooperative transportation tasks through narrow areas, whereas the method proposed in this paper shows significant improvements and better performance.

\section{Experiments and Analysis}

An outdoor experiment with three UAVs is conducted to validate the algorithm proposed in this paper. 
The UAV adopts a quadrotor configuration with a total weight of 1.47 kg. The propulsion system provides a maximum thrust of 30 N.
A fisheye camera is implemented for load localization. Inter-UAV communication via UDP protocol enables position sharing between agents.
The load consisted of a 3.06 kg box secured in a meshed enclosure. 
The video (URL: \url{https://youtu.be/_VUCs72NQ9s}) shows the experiment results at the end.

As depicted in the experimental results Fig.\ref{fig:expall}.(a) and (b), a preloaded virtual tube with a curved generating line and narrow sections guides three UAVs to cooperatively transport the load. During traversal through constricted zones, the UAV formation compressed laterally while autonomously restoring to its initial quasi-equilateral triangular configuration in expansive zones. The velocity profiles and cable tension distributions (shown in Fig.\ref{fig:expall}.(c) and (d)) indicate smooth UAV motion with physically consistent force allocation. Notably, this cooperative aerial transportation system successfully navigated geometrically complex environments without requiring explicit formation planning or manual intervention, demonstrating full autonomy in constrained space traversal.

\section{Conclusion}

This paper addresses the practical challenge of cooperative aerial transportation through narrow areas using multiple UAVs. Building upon virtual tube theory and dissipative cooperative principles, we propose a novel cooperative control framework. Large-scale simulations with ten UAVs and outdoor flight experiments with heavy loads validate that the proposed method exhibits three key advantages: (1) low computational demands, (2) scalability for large-scale swarms, and (3) operational safety/reliability. However, the current implementation requires a preplanned virtual tube. Future research will focus on autonomous virtual tube generation during flight through multisensor fusion (LiDAR/RGB-D cameras) to enable cooperative transportation in dynamically complex environments.







\bibliographystyle{Bibliography/IEEEtranTIE}
\bibliography{Bibliography/IEEEabrv,Bibliography/BIB_xx-TIE-xxxx}\ 

\begin{IEEEbiography}[{\includegraphics[width=1in,height=1.25in,clip,keepaspectratio]{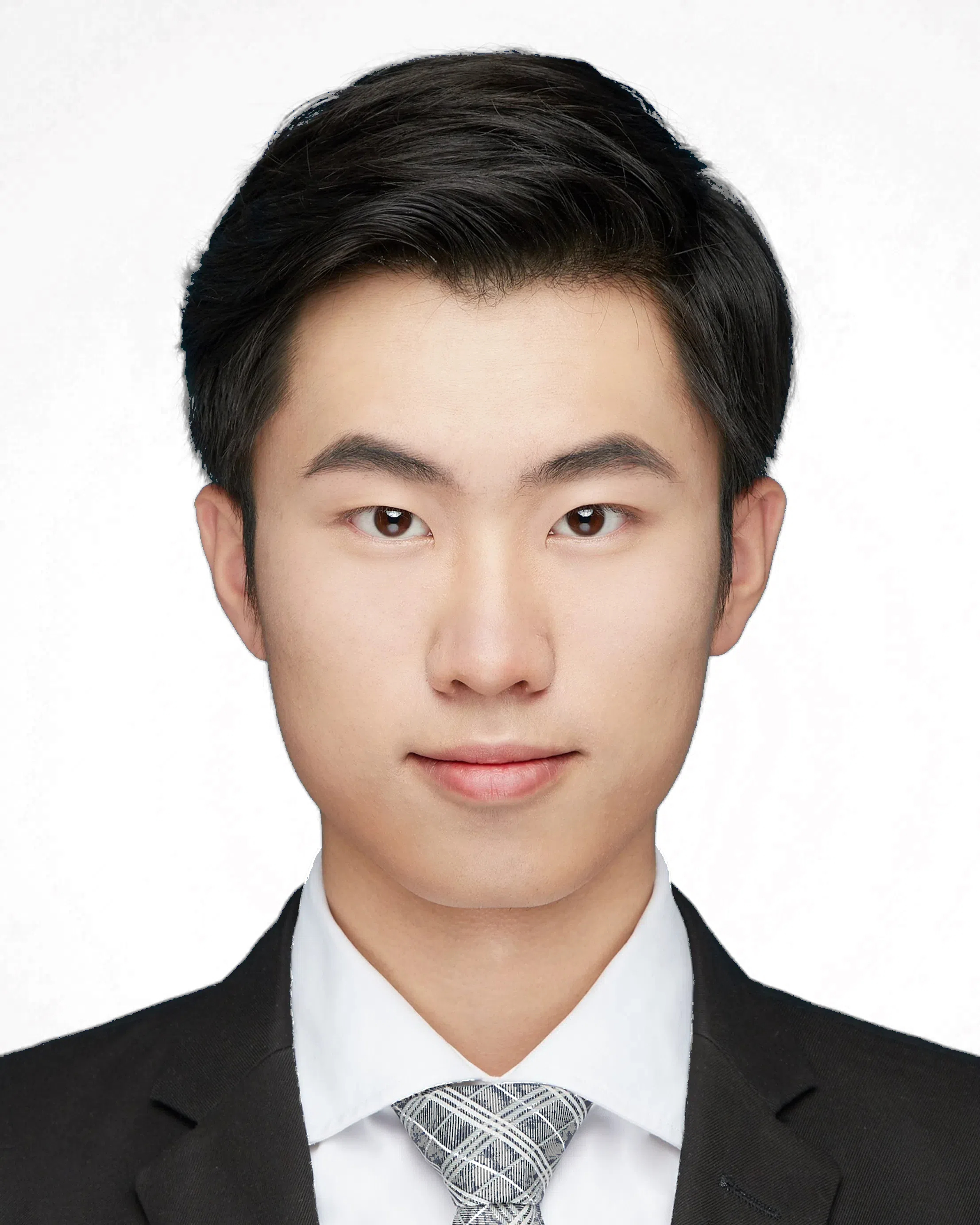}}]
	{Runxiao Liu} received the B.S. degrees in control science and engineering from Beihang University, Beijing, China, in 2021. He is currently working toward the Ph.D. degree in control science and
	engineering with the School of Automation Science and Electrical Engineering, Beihang University, Beijing, China.
	
	His research interests include multi-UAV cooperative transport control, image servo control.
\end{IEEEbiography}	
	
\vspace{-1cm}

\begin{IEEEbiography}[{\includegraphics[width=1in,height=1.25in,clip,keepaspectratio]{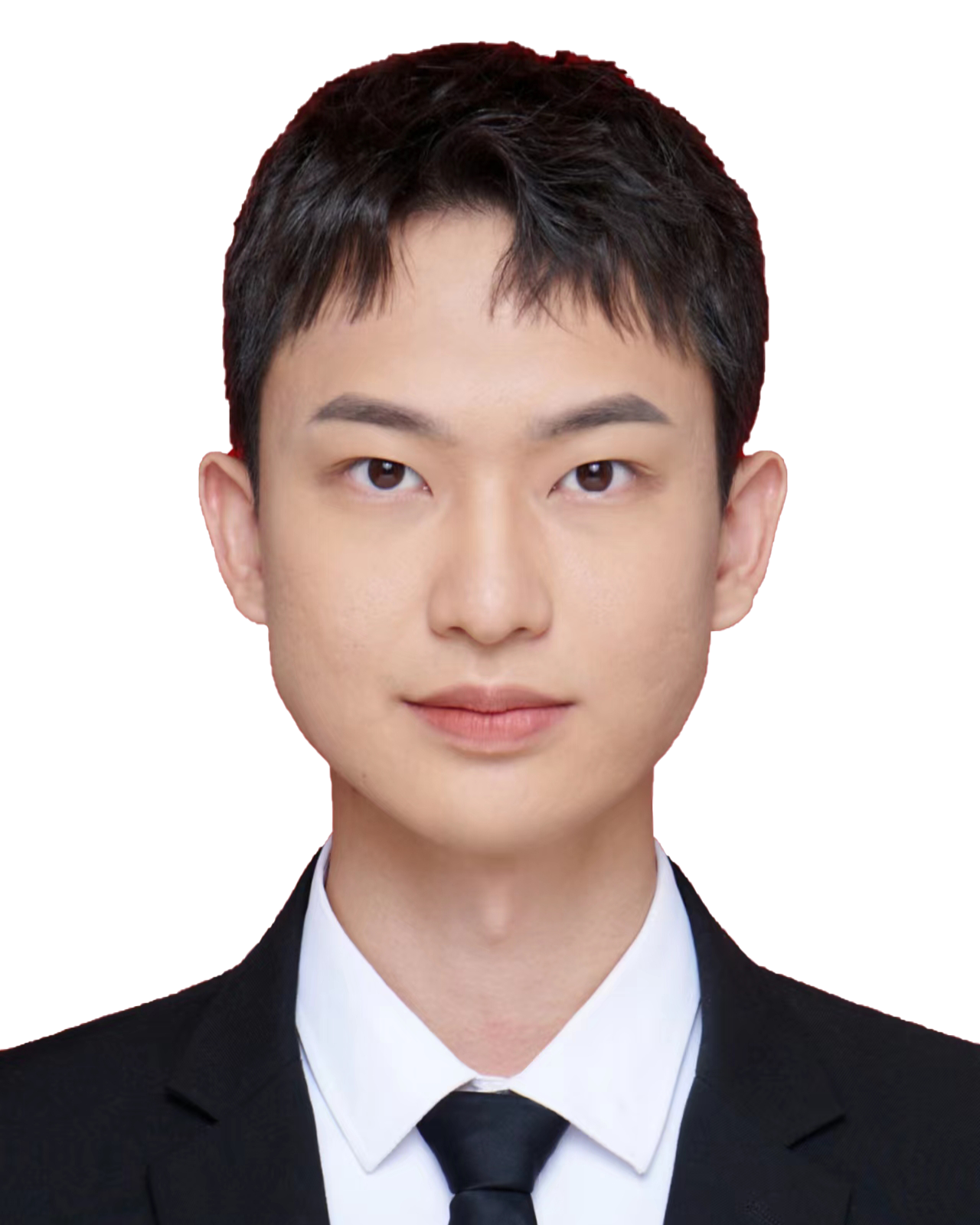}}]
	{Pengda Mao} received the B.S. degree in automation from Beihang University, Beijing, China, in 2020. He is currently working toward the Ph.D. degree in control science and engineering with the School of Automation Science and Electrical Engineering,Beihang University, Beijing, China. 
	
	His research interests include multi-UAV trajectory planning and control.
\end{IEEEbiography}

\vspace{-1cm}

\begin{IEEEbiography}[{\includegraphics[width=1in,height=1.25in,clip,keepaspectratio]{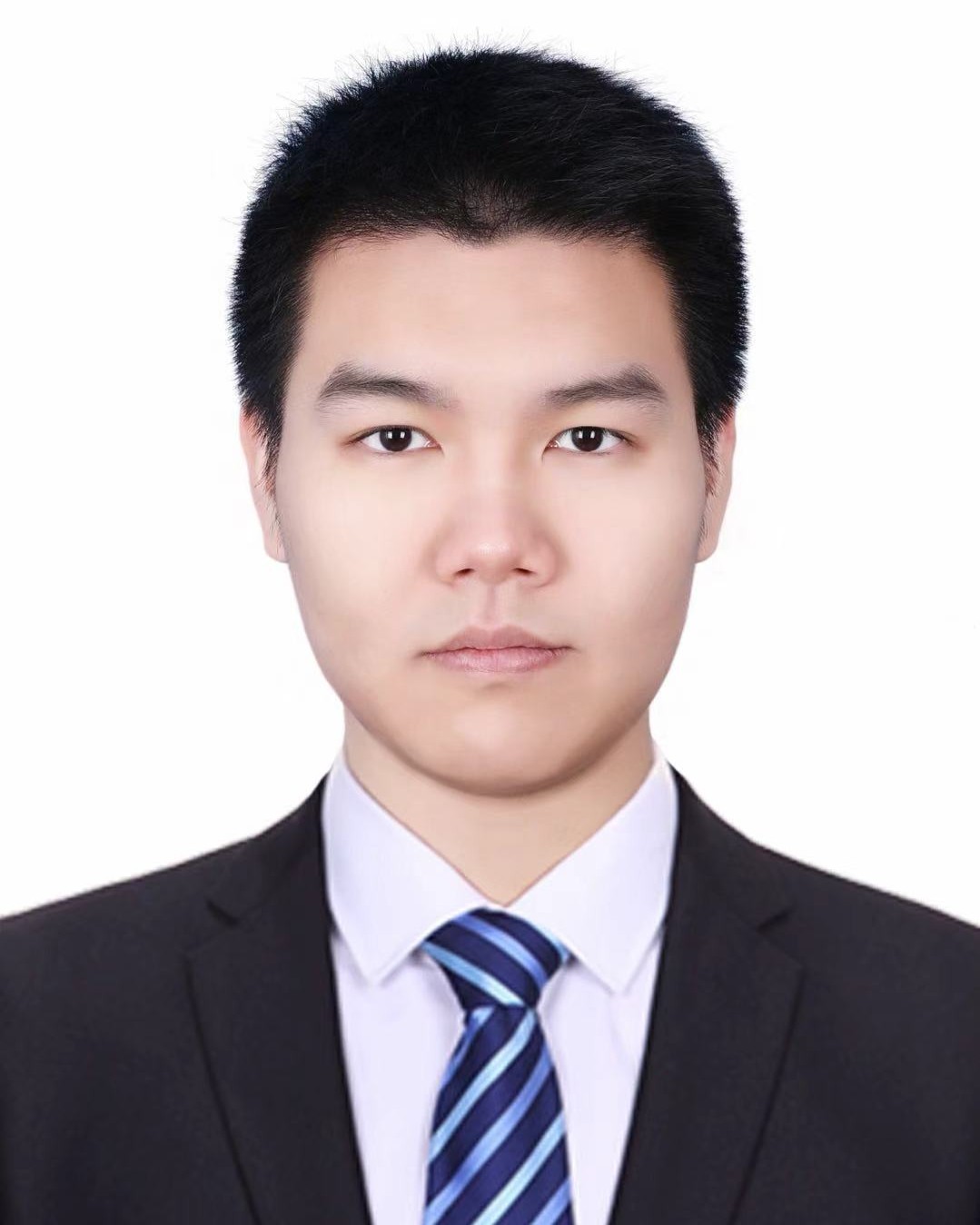}}]
	{Xiangli Le} received his B.S. degree in control science and engineering from Shenyuan Honors College, Beihang University, 2022. He is currently working toward the Ph.D. degree in control science and engineering with the School of Automation Science and Electrical Engineering, Beihang University, Beijing, China.
	
	His research interests include modular UAV transport, fault tolerant control and fault diagnosis.
\end{IEEEbiography}

\vspace{-1cm}

\begin{IEEEbiography}[{\includegraphics[width=1in,height=1.25in,clip,keepaspectratio]{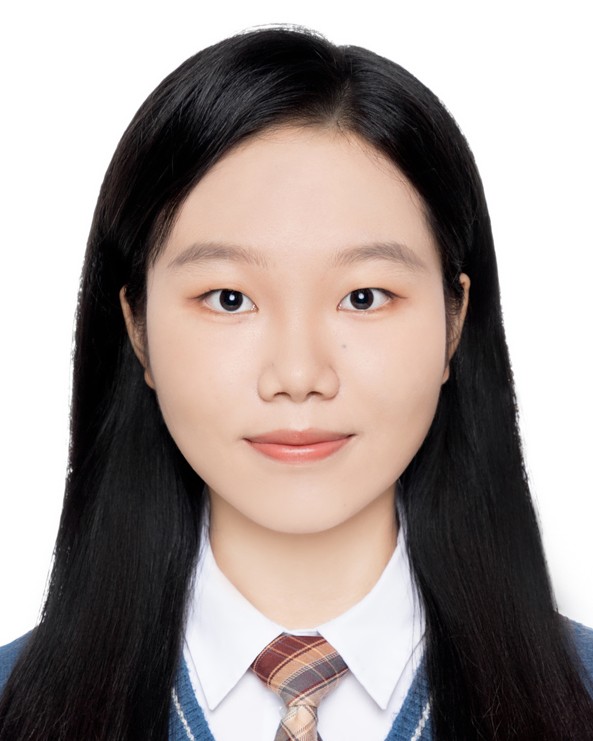}}]
{Shuang Gu} received the B.S. degrees in control science and engineering from Beihang University, Beijing, China, in 2024. She is currently working toward the Master's degree in control science and engineering with the School of Automation Science and Electrical Engineering, Beihang University ,Beijing ,China.

Her research interests include multi-UAV cooperative transport control, virtual tube.
\end{IEEEbiography}

\vspace{-1cm}

\begin{IEEEbiography}[{\includegraphics[width=1in,height=1.25in,clip,keepaspectratio]{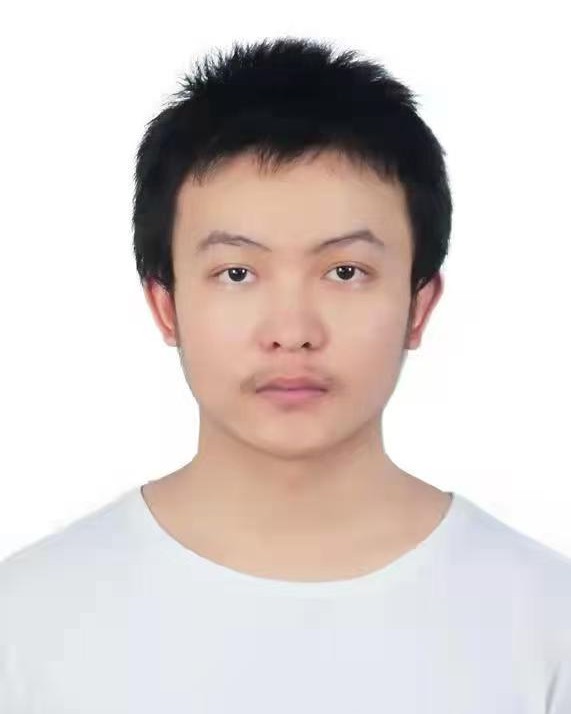}}]
{Chen Yapeng} is currently pursuing his Bachelor's degree in Robotics Engineering at Beihang University, Beijing, China. 

His research interests include multi-agent cooperative control.
\end{IEEEbiography}

\vspace{-1cm}
\begin{IEEEbiography}[{\includegraphics[width=1in,height=1.25in,clip,keepaspectratio]{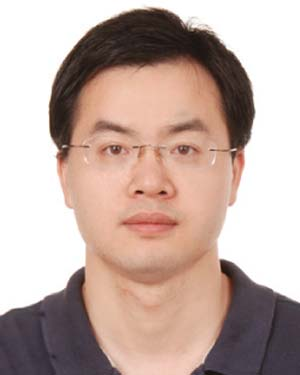}}]
{Quan Quan}(Senior Member, IEEE) received the B.S. and Ph.D. degrees in control science and engineering from Beihang University, Beijing, China, in 2004 and 2010, respectively.

He is a Professor with Beihang University, where he is currently with the School of Automation Science and Electrical Engineering. His research interests include reliable flight control, swarm intelligence, vision-based navigation, and health assessment.
\end{IEEEbiography}

\end{document}